\documentclass{article}
\usepackage{iclr2026_conference,times}
\usepackage{amsmath,amsfonts,bm}

\def\eqref#1{equation~\ref{#1}}

\def\1{\bm{1}}

\DeclareMathAlphabet{\mathsfit}{\encodingdefault}{\sfdefault}{m}{sl}
\SetMathAlphabet{\mathsfit}{bold}{\encodingdefault}{\sfdefault}{bx}{n}

\usepackage{booktabs}
\usepackage{multirow}
\usepackage{graphicx}
\usepackage{amsmath}
\usepackage{xcolor}
\usepackage{hyperref}
\usepackage{url}
\usepackage{microtype}
\usepackage{inconsolata}

\usepackage{algorithm}
\usepackage{algpseudocode}

\usepackage{marvosym}
\definecolor{linkblue}{HTML}{0052D9}
\hypersetup{
    breaklinks=true,
    colorlinks=true,
    citecolor=linkblue,
    linkcolor=linkblue,
    urlcolor=linkblue
}

\title{Environment Evolution for Terminal Agents}

\author{
\textbf{Zhiyuan Fan \quad Tinghao Yu \quad Yuanjun Cai \quad Jiang Zhou \quad Jiangtao Guan \quad Jincheng Liu}
\\[0.2cm]
\textbf{~Yun Yang \quad Dingxin Hu \quad Zhuo Han \quad Xing Wu \quad Feng Zhang \quad Lilin Wang}
\\[0.2cm]
Hunyuan Team, Tencent\\
\Letter~ zhiyuan.fan@connect.ust.hk,~\{maxwellyu, lilinwang\}@tencent.com\\
}

\iclrfinalcopy

\begin{document}
\maketitle

\begin{abstract}
Scaling interactive and verifiable environments is critical for training terminal agents.
As frontier models become more capable, environments synthesized from scratch become less challenging and thus provide limited learning signals. Recent co-evolution methods iteratively synthesize environments near the model's learnable frontier based on weaknesses exposed during rollouts.
However, their dependence on on-policy rollouts limits generalization and the continuous provision of learning signals as the model becomes stronger. In this paper, we propose \emph{environment evolution}, which incrementally increases environment difficulty off-policy and schedules the evolved environments generation by generation during training to provide continuous learning signals. We derive three evolution directions that influence environment difficulty from the multi-turn learning objective and then implement evolution along these directions through a loop-engineered multi-agent harness.
Quantitative rollout experiments with Hy4 preview, Claude Opus~5, and GPT-5.6 Sol show that environment evolution consistently produces more difficult environments.
We validate its effectiveness on Qwen3.6-27B and Qwen3.6-35B-A3B through simple long-horizon RL training, improving their performance by 14.4 and 18.0 percentage points on Terminal-Bench~2.1, respectively.
\end{abstract}

\section{Introduction}
Reinforcement learning environments are emerging as the next scalable direction for training capable agents~\citep{bellemare2013arcade,brockman2016openai}. With industrial-scale agentic RL algorithms stabilizing and asynchronous infrastructure maturing~\citep{fu2026areallargescaleasynchronousreinforcement,skyrlagent,earl}, the focus of further scaling is shifting toward the environments.
For general-purpose terminal agents, the difficulty and diversity of environments are therefore the central levers that determine what agents can learn through verifiable feedback: adaptive difficulty preserves continual learning potential, while diversity supports generality~\citep{dennis2021emergentcomplexityzeroshottransfer,jiang2021prioritizedlevelreplay,parkerholder2023evolvingcurricularegretbasedenvironment,garcin2024dred,cobbe2019procgen,openendedlearningteam2021openendedlearningleadsgenerally,merrill2026terminalbenchbenchmarkingagentshard}.

Recent work has focused on synthesizing large-scale terminal environments from scratch through human-designed pipelines that transform diverse resources (e.g., GitHub repositories, skills, and webpages) into executable terminal environments, each with a corresponding instruction (what to do) and a verification system (how completion is assessed)~\citep{gandhi2026endlessterminalsscalingrl,wu2026largescaleterminalagentictrajectory,fan2026scalableterminaltasksynthesis,pi2026dataengineeringscalingllm,hua2026cliuniverseverifiabletasksynthesis,zhao2026nexforge,yao2026learninggeneralizablebehaviorsterminal}.
However, these environments are often insufficiently challenging for current frontier models, which consistently solve them across repeated rollouts.
When used for RL training, such environments are discarded because they fail to provide effective learning signals that distinguish between better and worse trajectories, wasting environment construction costs and reducing diversity in the retained training distribution.

To provide useful learning signals, existing co-evolution methods couple model training with environment synthesis by using on-policy rollouts on seed environments to expose weaknesses that guide the synthesis of new environments that remain challenging yet learnable~\citep{zala2024envgengeneratingadaptingenvironments,hu2025agentgenenhancingplanningabilities,guo2025genenvdifficultyalignedcoevolutionllm,sygkounas2026covolveadversarialcoevolutionlargelanguagemodelgenerated}.
However, the resulting environments are constrained by the rollout model and initial environment distribution, limiting generalization and their ability to continuously provide learning signals as training saturates and model failures become sparse.

\begin{figure*}
    \centering
    \includegraphics[width=1\linewidth]{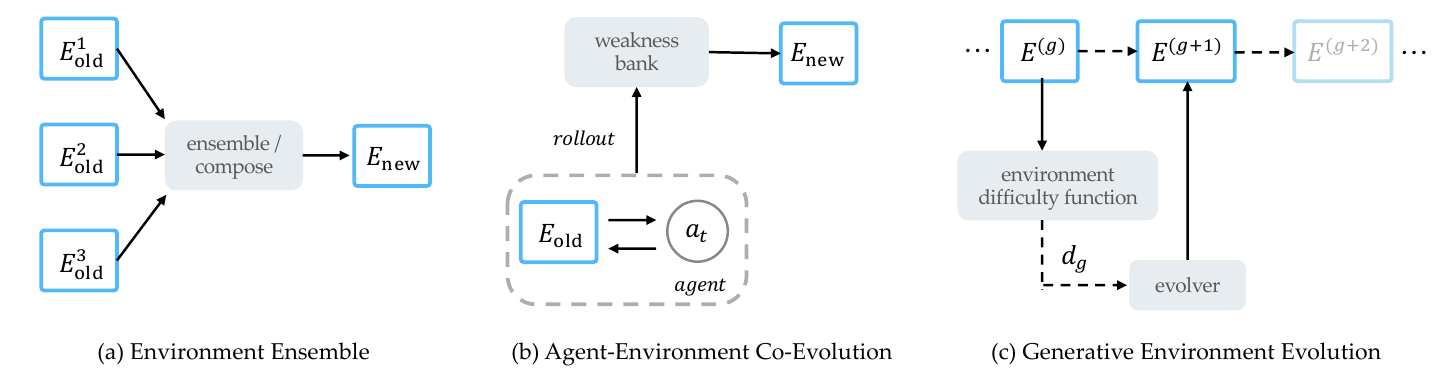}
    \caption{
    Comparison of paradigms for environment scaling.
    \textbf{(a)} Environment ensemble obtains a new environment by composing several difficult primitive environments.
    \textbf{(b)} Agent--environment co-evolution rolls out the target agent on primitive environments to populate a weakness bank, which guides
    the construction of new environments.
    \textbf{(c)} Environment evolution, our proposed paradigm, evolves the environment itself: an evolver uses
    a derived difficulty signal to produce successive environments along the lineage.
    }
    \label{fig:main}
\end{figure*}

In this paper, we propose \emph{environment evolution}, which increases environment difficulty generation by generation without relying on rollout models, as shown in Figure~\ref{fig:main}.

From the multi-turn learning objective, we derive an off-policy formulation of environment difficulty, which identifies \emph{scenario} novelty, \emph{skill} rarity, and execution \emph{length} as three factors that influence difficulty.
We then implement environment evolution with a loop-engineered multi-agent harness that incrementally modifies an existing environment along one selected from these three directions, which constructs lineages of increasing difficulty while introducing diverse variations around the original environment distribution.
Rollout-based difficulty estimates with Hy4 preview, Claude Opus~5, and GPT-5.6 Sol show that the evolved environments are challenging across different models despite being synthesised off-policy.
During model development, we find evolved environments continue to provide challenging RL signals even when their seed environments have already been used for SFT.

Experiments on Qwen3.6-27B and Qwen3.6-35B-A3B demonstrate that environment evolution improves Terminal-Bench~2.1 performance by 14.4 and 18.0 percentage points, respectively.
Compared with agent environment co-evolution, off-policy environment evolution provides longer-lasting learning signals and achieves better performance.
In summary, our contributions are as follows:

1. We derive a model-agnostic formulation of environment difficulty from the multi-turn learning objective, showing that environment evolution provides a more general approach to increasing environment difficulty.

2. We propose \emph{environment evolution} and implement it as a loop-engineered multi-agent harness that incrementally evolves existing environments to construct verified lineages of increasing difficulty.

3. We validate the approach through 200-step long-horizon RL experiments on Qwen3.6-27B and Qwen3.6-35B-A3B, demonstrating longer-lasting learning signals than co-evolution and ensemble baselines and improvements of 14.4 and 18.0 percentage points on Terminal-Bench~2.1, respectively.

\section{Related Work}

\paragraph{Terminal agents.}
Terminal agents interact with computing systems through command-line tools~\citep{chen2021evaluatinglargelanguagemodels,yang2024sweagentagentcomputerinterfacesenable,merrill2026terminalbenchbenchmarkingagentshard}, granting them open-ended access to explore and exploit computational resources and iteratively use environment feedback to complete long-horizon tasks~\citep{yao2023reactsynergizingreasoningacting}.
A line of work on harness design aims to strengthen agents' planning, navigation, and exploration while constraining undesirable behaviors, with a focus on observation-space design, context compression, and tool routing~\citep{lee2026metaharnessendtoendoptimizationmodel,wang2025openhandsopenplatformai,bui2026buildingeffectiveaicoding}.
The community has recently turned to synthesizing large-scale terminal environments from scratch for agent training~\citep{gandhi2026endlessterminalsscalingrl,wu2026largescaleterminalagentictrajectory,fan2026scalableterminaltasksynthesis,pi2026dataengineeringscalingllm,hua2026cliuniverseverifiabletasksynthesis,zhao2026nexforge,yao2026learninggeneralizablebehaviorsterminal}.
Despite their abundance and broad domain coverage, our large-scale rollout and quality assessment experiments reveal that these open-source environments suffer from low-quality reward signals (e.g., misalignment between task instructions and verification systems, corrupted environments)~\citep{bercovich2026makesgoodterminalagentbenchmark} and are insufficiently challenging to provide meaningful learning signals for frontier models.

\paragraph{Environment Scaling.}
Open-ended reinforcement learning requires a continual stream of solvable yet challenging environments that retain learning potential~\citep{wang2019pairedopenendedtrailblazerpoet,dennis2021emergentcomplexityzeroshottransfer,jiang2021prioritizedlevelreplay,parkerholder2023evolvingcurricularegretbasedenvironment}.
Paired Open-Ended Trailblazer (POET) co-evolves a population of environment--agent pairs, generating new challenges through environment mutation and transferring agents across environments to exploit stepping stones~\citep{wang2019pairedopenendedtrailblazerpoet,wang2020enhancedpoetopenendedreinforcement}.
Unsupervised Environment Design (UED) formalizes the automatic construction and curation of valid and solvable environments from underspecified environment parameters, encompassing regret-based generation, prioritized replay, and incremental level editing~\citep{dennis2021emergentcomplexityzeroshottransfer,jiang2021prioritizedlevelreplay,jiang2022replayguidedadversarialenvironmentdesign,parkerholder2023evolvingcurricularegretbasedenvironment}.
Recent work has begun to bring these ideas to LLM agents: through feedback-conditioned generation~\citep{chen2025scalingagentlearning,yang2026coevolve}, online curricula~\citep{qi2025webrltrainingllmweb}, and agent--environment co-evolution~\citep{guo2025genenvdifficultyalignedcoevolutionllm,liu2026spade}, these methods adapt the environment distribution as the agent improves, keeping environments near its capability frontier.
However, they require a designated agent to estimate environment difficulty through on-policy rollouts, and the resulting environments are related to both the rollout agent and the initial environment distribution.
Instead, environment evolution constructs increasingly difficult lineages independently of the target policy and schedules successive generations during training to provide continual learning signals.

\section{Preliminaries}
\label{sec:preliminaries}

Instead of estimating difficulty on-policy, e.g., by rolling out a model and using its pass rate as the difficulty metrics, we need an off-policy metric that measures the difficulty of the environment itself, derived from the multi-turn learning objective. Since models are trained on different data distributions, a difficulty estimate tied to one model \(\theta\) is model-specific weakness rather than environment difficulty.

We first view agent-environment interaction as a Markov process~\citep{kaelbling1998planning} with interleaved observations and actions.
Let \(h_t=(o_{\le t},a_{<t})\).
Then \(o_t\sim O_{\mathcal E}(\cdot\mid s_t)\), \(a_t\sim\pi_\theta(\cdot\mid h_t,g)\), and \(s_{t+1}\sim P_{\mathcal E}(\cdot\mid s_t,a_t)\), which induces a low-level execution trajectory \(\zeta=(o_0,a_0,o_1,a_1,\ldots,o_T)\).

Following prior definitions from hierarchical agent execution \citep{sutton1999between}, we treat agent execution trajectory at a higher level as an interleaving of scenarios and skill executions:
\[
\xi=(\sigma_0,\kappa_1,\sigma_1,\ldots,\kappa_L,\sigma_L),
\]
where \(\sigma_t\) is the high-level scenario at step \(t\), and \(\kappa_t\) is the skill applied under that scenario.

Under model \(\theta\), the likelihood of a high-level trajectory decomposes over the scenarios it reaches and the skills it applies.
Taking the negative log-likelihood gives the model-specific difficulty:
\[
\begin{aligned}
D_{\theta}(\xi)
&= -\log p_\theta(\xi\mid g)\\
&= \sum_{t=1}^{L} \Bigl(
-\log p_\theta(\sigma_{t-1}\mid g)
-\log p_\theta(\kappa_t\mid\sigma_{t-1},g)
\Bigr).
\end{aligned}
\]

This quantity has three contributors.
First, \(L\) is the number of meaningful solver turns required by the trajectory.
Second, \(-\log p_\theta(\sigma_{t-1}\mid g)\) measures scenario novelty under the model.
Third, \(-\log p_\theta(\kappa_t\mid\sigma_{t-1},g)\) measures the rarity of applying the required skill in that scenario.
The last two terms are policy-dependent: they depend on the model's training-data distribution and learned policy.

To obtain a policy-independent difficulty measure, we replace the model-dependent probabilities with those under a reference distribution \(\mathcal T\) grounded in broad world knowledge:
\[
D_{\mathcal T}(\xi)
= \sum_{t=1}^{L} \Bigl(
-\log p_{\mathcal T}(\sigma_{t-1}\mid g)
-\log p_{\mathcal T}(\kappa_t\mid\sigma_{t-1},g)
\Bigr).
\]
Here \(p_{\mathcal T}(\sigma\mid g)\) measures how common a scenario is within the environment family, and \(p_{\mathcal T}(\kappa\mid\sigma,g)\) measures how common the required skill is under that scenario.
This converts the estimate from model-specific weakness into model-agnostic environment difficulty.
In particular, a deep-research agent can estimate both distributions through broad web search, grounding them in world knowledge, by providing the relative context of \(g\) and \(\xi\).

This also gives a direct way to relate environment difficulty to agent weakness.
Let \(z_t=(\sigma_{t-1},\kappa_t)\) denote the scenario-skill requirement at step \(t\), and define the per-step difficulties
\[
d_\theta(z_t\mid g)
= -\log p_\theta(\sigma_{t-1}\mid g)
-\log p_\theta(\kappa_t\mid\sigma_{t-1},g),
\]
and \(d_{\mathcal T}(z_t\mid g)\) analogously under \(p_{\mathcal T}\).
Agent weakness is the excess difficulty that remains after subtracting the environment-family difficulty:
\[
\delta_\theta(z_t\mid g)
=
\left[
d_\theta(z_t\mid g)
-
d_{\mathcal T}(z_t\mid g)
\right]_+ .
\]
where \([x]_+=\max(x,0)\).
Equivalently,
\[
\delta_\theta(z_t\mid g)
=
\Bigl[
\log
\frac{p_{\mathcal T}(\sigma_{t-1}\mid g)}
     {p_\theta(\sigma_{t-1}\mid g)}
+
\log
\frac{p_{\mathcal T}(\kappa_t\mid\sigma_{t-1},g)}
     {p_\theta(\kappa_t\mid\sigma_{t-1},g)}
\Bigr]_+ .
\]
For the full high-level trajectory,
\[
\Delta_\theta(\xi)
=
\sum_{t=1}^{L}
\left[
d_\theta(z_t\mid g)
-
d_{\mathcal T}(z_t\mid g)
\right]_+ .
\]
Thus, weakness is not the difficulty of the environment itself; it is the portion of that trajectory that is unusually difficult for a particular model relative to the environment family.

This distinction clarifies the scope of on-policy co-evolution of agent and environment.
Such methods collect rollouts from a model \(\theta\), identify the model's failure modes, and generate new environments around those failures.
If \(e_t^\theta\) indicates a failure at step \(t\), the induced signal is mainly \(\sum_{t=1}^{L} e_t^\theta[-\log p_\theta(\kappa_t\mid\sigma_{t-1},g)]\).
That is, on-policy co-evolution of agent and environment primarily targets skill-selection errors under scenarios contained in the seed environments and reached by the current model during rollouts.
It does not explicitly control the number of required solver steps \(L\), nor does it systematically increase scenario novelty under \(p_{\mathcal T}(\sigma\mid g)\).
In contrast, the environment evolution directly operates on the full difficulty space:
\[
\left(
L,\;
-\log p_{\mathcal T}(\sigma\mid g),\;
-\log p_{\mathcal T}(\kappa\mid\sigma,g)
\right)
\]
which shows that environment evolution offers a more general paradigm for providing continuous learning signals.

\section{Approach}

\subsection{Sequence-Guided Environment Evolution.}
\begin{figure}[H]
    \centering
    \includegraphics[width=\linewidth]{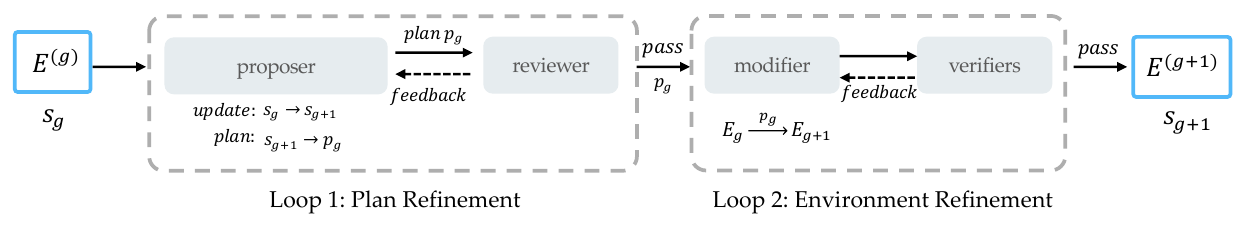}
    \caption{Loop-engineered multi-agent harness for environment evolution. It decomposes each generation into two gated feedback loops: (1) sequence-guided plan refinement, which generates and revises an evolution plan until it passes rubric-based review; and (2) plan-conditioned environment refinement, which evolves and repairs the candidate until strict solvability and quality checks pass.}
    \label{fig:environment-evolution-harness}
\end{figure}

Environment evolution is implemented as a loop-engineered multi-agent harness,
as illustrated in Figure~\ref{fig:environment-evolution-harness}. It takes the
latest accepted environment as input and produces the next-generation
environment through an incremental modification guided by the expected
execution trajectory at the scenario and skill levels.

\paragraph{Loop 1: Plan Refinement}
The Proposer first extracts an execution sequence of interleaved scenarios and
skills from \(E\):
\begin{equation}
\xi_E
=
(\sigma_1,\kappa_1),\ldots,(\sigma_L,\kappa_L).
\end{equation}
It then updates the sequence according to the evolution direction selected at
the current generation. For \emph{length}, it inserts scenario--skill pairs
into the sequence to introduce additional dependencies along the expected
execution trajectory. For \emph{scenario}, it replaces one scenario while preserving
its paired skill; for \emph{skill}, it replaces one skill.
A plan is generated from the difference between the updated and
original sequences, and a rubric-based reviewer iteratively reviews it until it
is accepted or the current evolution direction fails.

\paragraph{Loop 2: Environment Refinement}
The reviewed plan is then passed to the Modifier, which creates a residual
\(\Delta E\) and applies it to the current environment. Each candidate must pass
three verifiers run in parallel: (i) an Oracle verifier checks that the reference
solution succeeds in the sandbox; (ii) an Invalid-test verifier confirms that an
empty or no-op solution fails, ensuring that the verification system is reliable;
and (iii) an adaptive general-rubrics verifier checks environment quality.
During development, accepted environments undergo human-in-the-loop review.
Issues that bypass the rubric-based checks are converted into new rubrics
for the plan reviewer and environment verifier until the loop reliably produces
environments with no issues identified by human reviewers.

\paragraph{Evolution effort.}
To control the mutation between adjacent generations, we introduce a
prompt-controlled mutation parameter called \emph{evolution effort}, analogous
in spirit to thinking effort, with three levels: low, high, and max. While the
evolution direction determines what type of sequence-level edit is performed,
evolution effort controls its scope. The three effort levels restrict the edit
to one pair \((\sigma_\ell,\kappa_\ell)\), one contiguous span, or an unrestricted portion of
the sequence, respectively.
Section~\ref{sec:experiments} quantitatively validates the effectiveness of
this design.

At each generation, we randomly order the three evolution directions. If the
Plan Reviewer rejects a plan or the repair budget for the current direction
is exhausted, we fall back to the next direction, construct a new target
sequence from the same environment, and repeat the process. The branch
terminates only when all three directions fail.

\subsection{Evolution-Lineage Scheduler.}
As evolution proceeds, later generations become increasingly difficult.
Sampling randomly from the full lineage can therefore expose the policy to
environments it cannot yet solve, yielding all-failure rollout groups that
provide no effective learning signal. We therefore propose the
Evolution-Lineage (EL) Scheduler, which starts from the earliest generation and
schedules its environments in order. Let \(E_{i,g,k}\) be the \(k\)-th
environment in generation \(g\) of lineage \(i\), with \(N_{i,g}\) environments
in that generation. At update \(u\), the scheduler updates the active indices as
follows:
\begin{equation}
\label{eq:environment-scheduler}
(g_{u+1},k_{u+1})
=
\begin{cases}
(g_u,k_u), & \widehat p_u(E_{i,g_u,k_u})\leq\tau,\\
(g_u,k_u+1), & \widehat p_u(E_{i,g_u,k_u})>\tau
    \ \land\ k_u<N_{i,g_u},\\
(g_u+1,1), & \widehat p_u(E_{i,g_u,k_u})>\tau
    \ \land\ k_u=N_{i,g_u},
\end{cases}
\qquad
\widehat p_u(E)=\frac{1}{B}\sum_{b=1}^{B}r_{u,b}.
\end{equation}
We use \(B=8\) and \(\tau=6/8\). Once the current environment exceeds this
threshold, the scheduler moves to the next environment in the same generation.
It advances to the next generation only when no environments remain in the
current one.

\section{Experiments}
\label{sec:experiments}

\subsection{Experimental Setup}

\paragraph{Harness.}
In this paper, evaluation and RL training use the Claude Code harness~\citep{anthropic2025claudecode}.
It fixes the tool protocol across models while improving execution efficiency by running multiple tool calls in parallel within a single assistant turn.
For RL training, the harness operates with a 256K context window and automatically compacts the trajectory when the remaining usable context reaches 16K, providing stable context management for long-horizon execution.

\paragraph{Benchmark.}
Terminal-Bench~2.1 Verified~\citep{merrill2026terminalbenchbenchmarkingagentshard} is the primary held-out benchmark for terminal agent capability.
It fixes instabilities in Terminal-Bench~2.0 that hinder reproducible evaluation and incorrectly underestimate benchmark performance.
Each task is configured with 32 CPU cores and 48\,GB of memory, with a timeout of 4 hours.
Sampling uses temperature \(1.0\), top-\(p\) \(0.95\), and top-\(k\) \(20\), with a dynamic output budget, \(256\mathrm{K}-L_{\mathrm{used}}\), at each assistant turn, to avoid truncation errors caused by an overly long turn.
We report the average over five runs.

\paragraph{Training Algorithm.}

We use GRPO~\citep{shao2024deepseekmathpushinglimitsmathematical} for agentic RL training with partial rollouts, fully asynchronous GPU location, and a staleness bound of 5 to reduce GPU bubble time.
When auto compact of Claude Code harness is triggered, its summary is retained in the complete trajectory and treated as a regular action turn for multi-turn credit assignment.
We train Qwen3.6-27B, a 27B-parameter dense model, and Qwen3.6-35B-A3B, a mixture-of-experts model with 35B total and 3B activated parameters.
Both checkpoints provide a native context length of 262,144 tokens. For the MoE model, we additionally use R3~\citep{ma2025stabilizingmoereinforcement} to stabilize training process.

\paragraph{Monitoring Metrics.}
\textbf{Environment Difficulty} is estimated by the pass rate over 8 independent rollouts.
We also record the average number of assistant turns as a measure of long-horizon execution; assistant turns typically account for nearly half of the total turns.
Rollouts use Claude Opus~5 and GPT-5.6 Sol with \texttt{xhigh} effort and Hy4 preview~\citep{tencent2026hy4preview} with \texttt{high} effort, all with a 1M context window.
\textbf{Environment Mutation.}
For each generation, structural change relative to the seed is measured across instruction, environment, and verification system, with their mutations measured at the token, file, and test-unit levels, respectively.

\paragraph{Seed Environment Selection.}
We collect 47,678 non-benchmark terminal environments from Hugging Face and GitHub and retain 127 through strict rubric-based filtering for environment quality and solvability.
Each candidate must include an executable Oracle solution that passes the verifier, satisfy rubric-based quality checks (we find environment quality critical to successful RL training), and meet the difficulty threshold under Claude Opus~5: a pass rate of at most \(4/8\) and average turns of at least 30.
SkillSynth~\citep{fan2026scalableterminaltasksynthesis} is then used to supplement the pool with newly synthesized environments, which are subject to the same quality and difficulty filters.
Finally, uniform sampling across domains yields a balanced and diverse seed pool of 500 environments.

\subsection{Evolution Effort}

\begin{figure}[H]
    \centering
    \includegraphics[width=\linewidth]{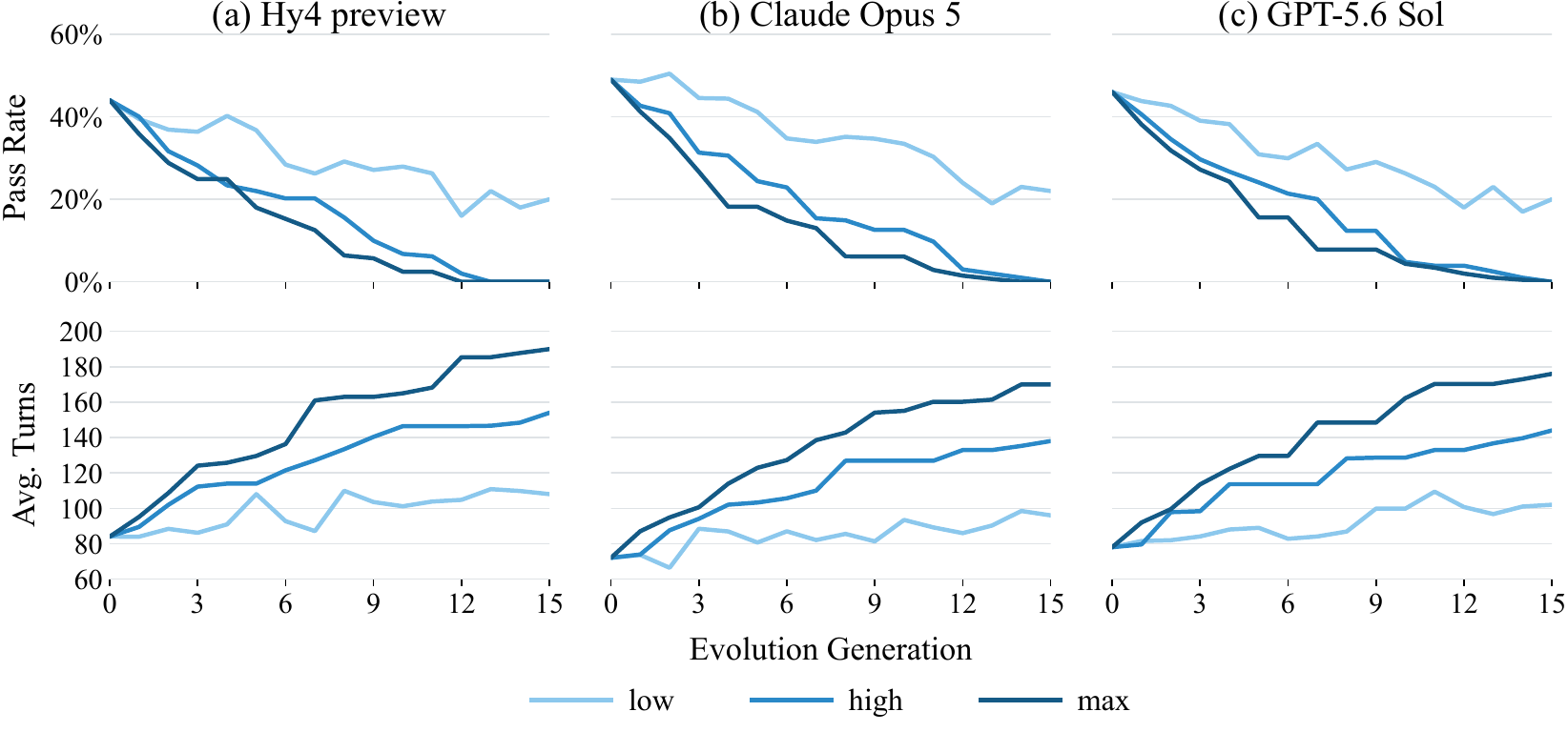}
    \caption{Environment difficulty across 15 evolution generations under \texttt{low}, \texttt{high}, and \texttt{max} evolution effort, evaluated independently by Hy4 preview, Claude Opus~5, and GPT-5.6 Sol.}
    \label{fig:evolution-effort-difficulty}
\end{figure}

Starting from the same seed environments, the randomized cross-mode strategy independently constructs 15-generation lineages under \texttt{low}, \texttt{high}, and \texttt{max} evolution effort.
We use a common cap of 15 generations because pass rate provides no further resolution once a lineage enters the zero-pass regime; beyond that point, the effectiveness of additional evolution cannot be assessed reliably from rollout outcomes.

As shown in Figure~\ref{fig:evolution-effort-difficulty}, \texttt{low} effort progressively extends long horizon execution, as avg turns increase across generations, but its pass rate fluctuates and remains above zero.
Because \texttt{low} modifies only one local pair \((\sigma_\ell,\kappa_\ell)\), successive generations may repeatedly edit the same pair and partially return to an earlier configuration.
In contrast, \texttt{high} and \texttt{max} monotonically reduce pass rate to zero and increase avg turns, with \texttt{max} reaching zero earlier and producing the larger change in difficulty.

\begin{figure}[H]
    \centering
    \includegraphics[width=\linewidth]{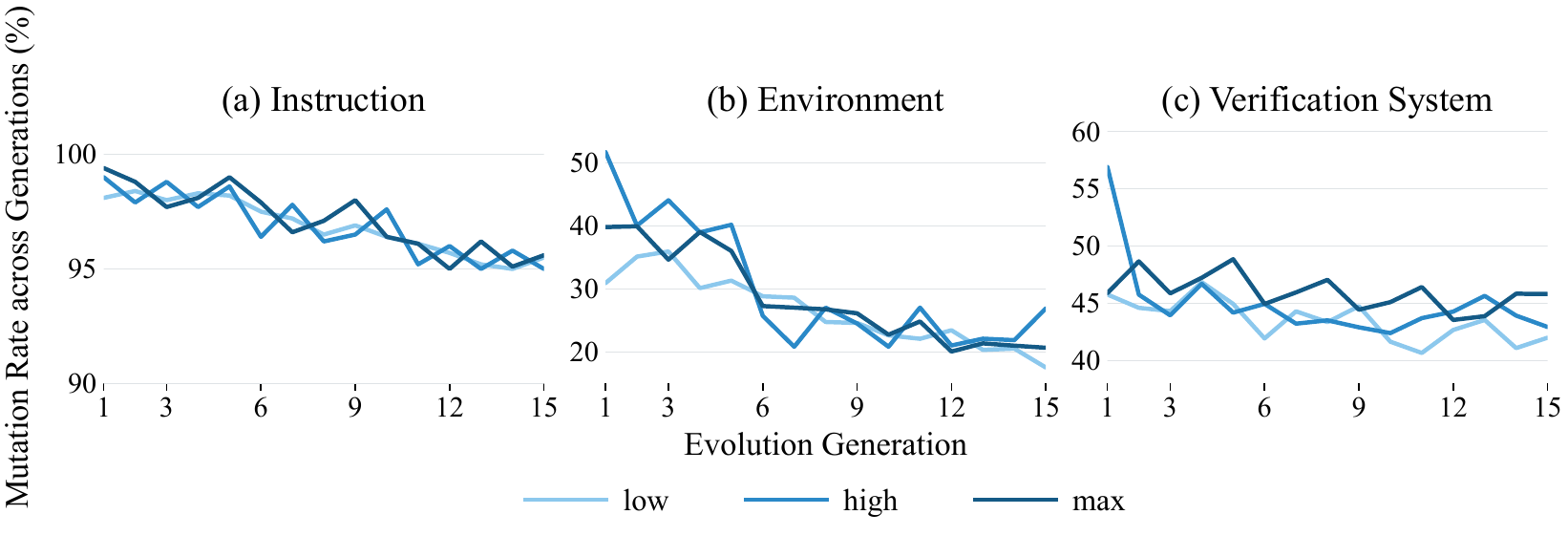}
    \caption{Instruction, environment, and verification system mutation rates across generations G1--G15 under \texttt{low}, \texttt{high}, and \texttt{max} evolution effort.}
    \label{fig:evolution-effort-mutation}
\end{figure}

Figure~\ref{fig:evolution-effort-mutation} separates generation-level mutation rates into instruction, environment, and verification system components.
Instruction mutation remains high under \texttt{high} and \texttt{max}, with both fluctuating between 95\% and 100\% while trending downward, whereas environment and verification system changes remain more selective.
Balancing effective and stable evolution against the controllability of each generation, \texttt{high} is used as the default evolution effort.

\subsection{Evolution Direction}

\begin{table}[H]
\centering
\small
\setlength{\tabcolsep}{8pt}
\renewcommand{\arraystretch}{1.12}
\begin{tabular}{lcccccc}
\toprule
\multicolumn{1}{c}{\multirow{2}{*}{Direction}}
& \multicolumn{2}{c}{Difficulty Change}
& \multicolumn{4}{c}{Component Mutation (\%)} \\
\cmidrule(lr){2-3}\cmidrule(lr){4-7}
& \(\Delta\) Pass Rate
& \(\Delta\) Avg Turns
& Instruction
& Environment
& Verification
& Total \\
\midrule
\multicolumn{7}{c}{1-step effect} \\
scenario & \(-4.7\,\mathrm{pp}\) & \(+13.5\) & \(99.8\%\) & \(59.6\%\) & \(53.8\%\) & \(71.1\%\) \\
skill    & \(-4.0\,\mathrm{pp}\) & \(+12.5\) & \(99.1\%\) & \(50.3\%\) & \(58.4\%\) & \(69.3\%\) \\
length   & \(-7.1\,\mathrm{pp}\) & \(+9.4\)  & \(87.5\%\) & \(45.2\%\) & \(58.6\%\) & \(63.8\%\) \\
\addlinespace
\multicolumn{7}{c}{15-step mean effect} \\
scenario & \(-2.9\,\mathrm{pp}\) & \(+9.5\)  & \(97.1\%\) & \(34.6\%\) & \(43.7\%\) & \(58.5\%\) \\
skill    & \(-2.7\,\mathrm{pp}\) & \(+10.6\) & \(96.5\%\) & \(27.8\%\) & \(45.8\%\) & \(56.7\%\) \\
length   & \(-4.8\,\mathrm{pp}\) & \(+7.4\)  & \(85.5\%\) & \(34.3\%\) & \(48.9\%\) & \(56.2\%\) \\
\bottomrule
\end{tabular}
\caption{Single-step consistency of evolution-direction effects under \texttt{high} evolution effort. The two difficulty-change columns are measured by Claude Opus~5 with \texttt{xhigh} thinking effort. The first block applies each direction once to the same seed environments. The second evolves each direction for 15 consecutive steps and averages the metrics over the 15 generation transitions. Agreement between the two blocks tests whether each direction preserves a stable, direction-specific change profile along longer lineages. Difficulty changes are computed as the later generation minus the previous generation, component columns report generation-level mutation rates, and Total is their unweighted mean.}
\label{tab:evolution-direction}
\end{table}

To isolate the effect of evolution direction, we fix \texttt{high} evolution effort and apply \texttt{scenario}, \texttt{skill}, and \texttt{length} to the same seed environments.
At each generation, we measure changes in pass rate and avg turns together with instruction, environment, and verification system mutation.
The 1-step effect captures the immediate change induced by each direction, while the 15-step mean averages the same metrics over adjacent transitions in a direction-specific lineage to test whether that effect persists.

All three directions consistently reduce pass rate and increase avg turns.
\texttt{length} produces the largest pass-rate decrease, whereas \texttt{scenario} and \texttt{skill} produce larger increases in avg turns.
\texttt{scenario} yields the largest total mutation, while \texttt{length} achieves the strongest pass-rate reduction with the smallest total mutation.
The agreement between the 1-step and 15-step results indicates that these direction-specific profiles persist beyond a single edit.
In the full evolution procedure, the directions are therefore randomly ordered at each generation to diversify the resulting lineages, with cross-mode fallback whenever the current direction fails.

\subsection{RL Training Dynamics}
Before RL training, each base model generates trajectories for rejection sampling fine-tuning (RFT)~\citep{touvron2023llama2}.
Only trajectories that pass the verification system and conform to the Claude Code protocol are retained.
The filtered set is rebalanced for trajectory diversity in order to increase policy entropy for better RL training.
Starting from the resulting RFT checkpoint, denoted step 0, we train each model with GRPO for 200 steps.

\begin{figure}[H]
    \centering
    \includegraphics[width=\linewidth]{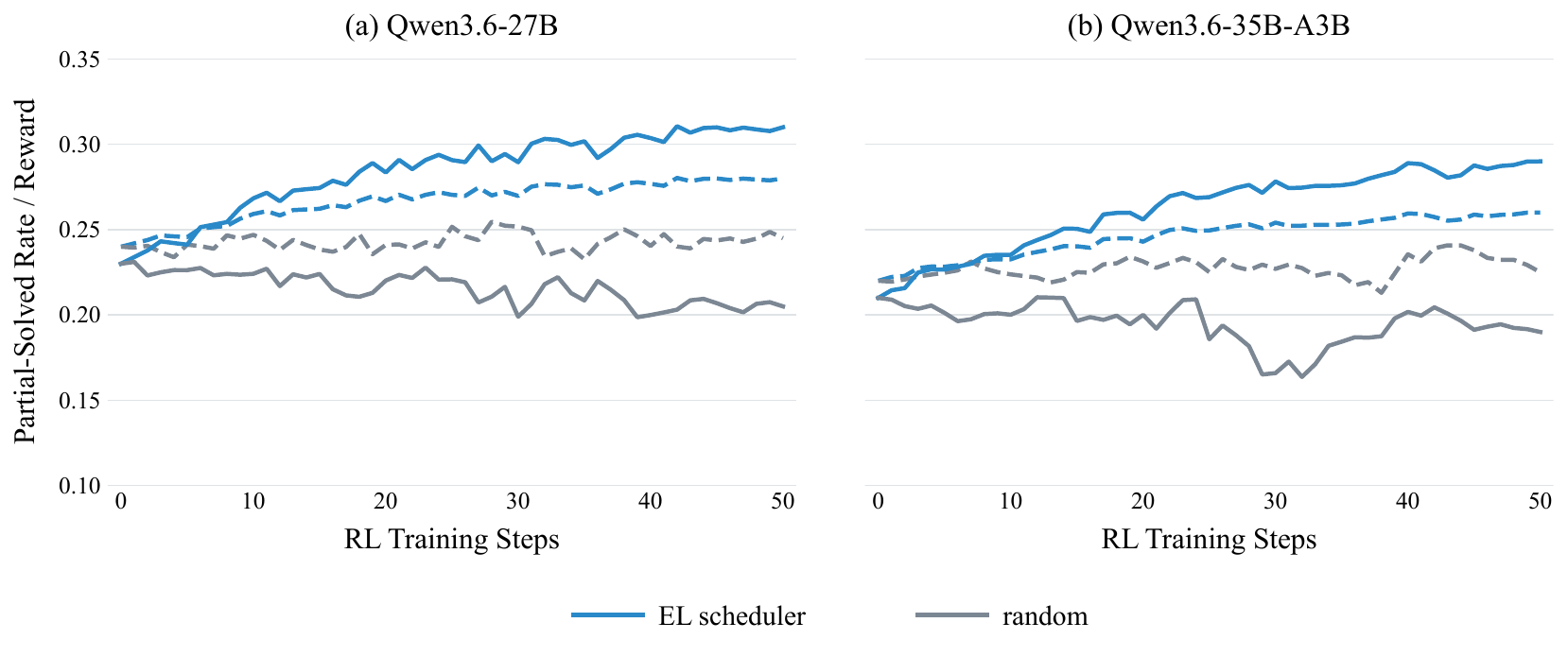}
    \caption{Early training dynamics over the first 50 training steps. Solid curves show the fraction of rollout groups that are partially solved, with one to seven successful trajectories among eight rollouts; dashed curves show mean training reward.}
    \label{fig:scheduler-dynamics}
\end{figure}

\paragraph{EL Scheduler.}
The EL Scheduler advances through each environment lineage in order, exposing generation \(g+1\) only after generation \(g\) reaches the pass-rate threshold in Equation~\ref{eq:environment-scheduler}.
Figure~\ref{fig:scheduler-dynamics} compares it with random scheduling under matched rollout budgets during the first 50 RL steps.
By avoiding premature exposure to generations that the policy cannot yet solve, the scheduler uses the rollout budget more efficiently and provides GRPO with more informative learning signals, as more rollout groups remain partially solved and retain non-zero within-group advantage.
The EL Scheduler is used by default throughout the full 200-step RL training run.

\begin{figure}[H]
    \centering
    \includegraphics[width=\linewidth]{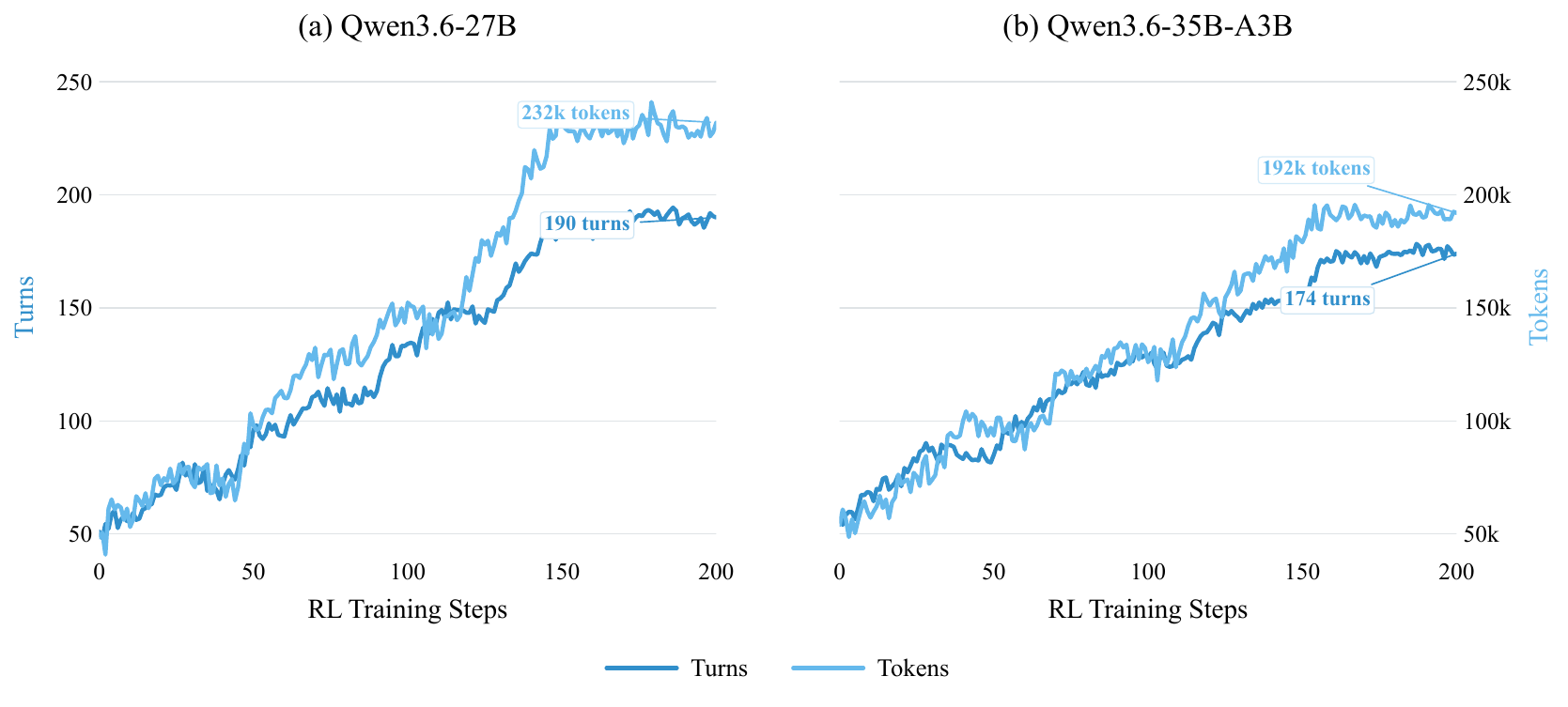}
    \caption{Trajectory length over 200 RL training steps for Qwen3.6-27B (left) and Qwen3.6-35B-A3B (right); the left and right y-axes report average turns and tokens per trajectory, respectively, with annotations at step 200. Tokens per turn increase from approximately 951 to 1,221 for Qwen3.6-27B and from 947 to 1,103 for Qwen3.6-35B-A3B.}
    \label{fig:rollout-training-dynamics}
\end{figure}

\paragraph{Turns and Tokens.}
During training, the scheduler progressively admits harder environments, which tend to require longer interaction horizons.
As shown in Figure~\ref{fig:rollout-training-dynamics}, this progression is accompanied by increases in both turns and tokens per trajectory.
Tokens per turn also increase, suggesting that the policy allocates greater thinking effort to each interaction.

\subsection{Comparison}
To ensure a fair comparison across environment-scaling paradigms, Claude Opus~5 is fixed as the environment synthesis model, and the total number of training environments is held constant.

\begin{figure}[H]
    \centering
    \includegraphics[width=\linewidth]{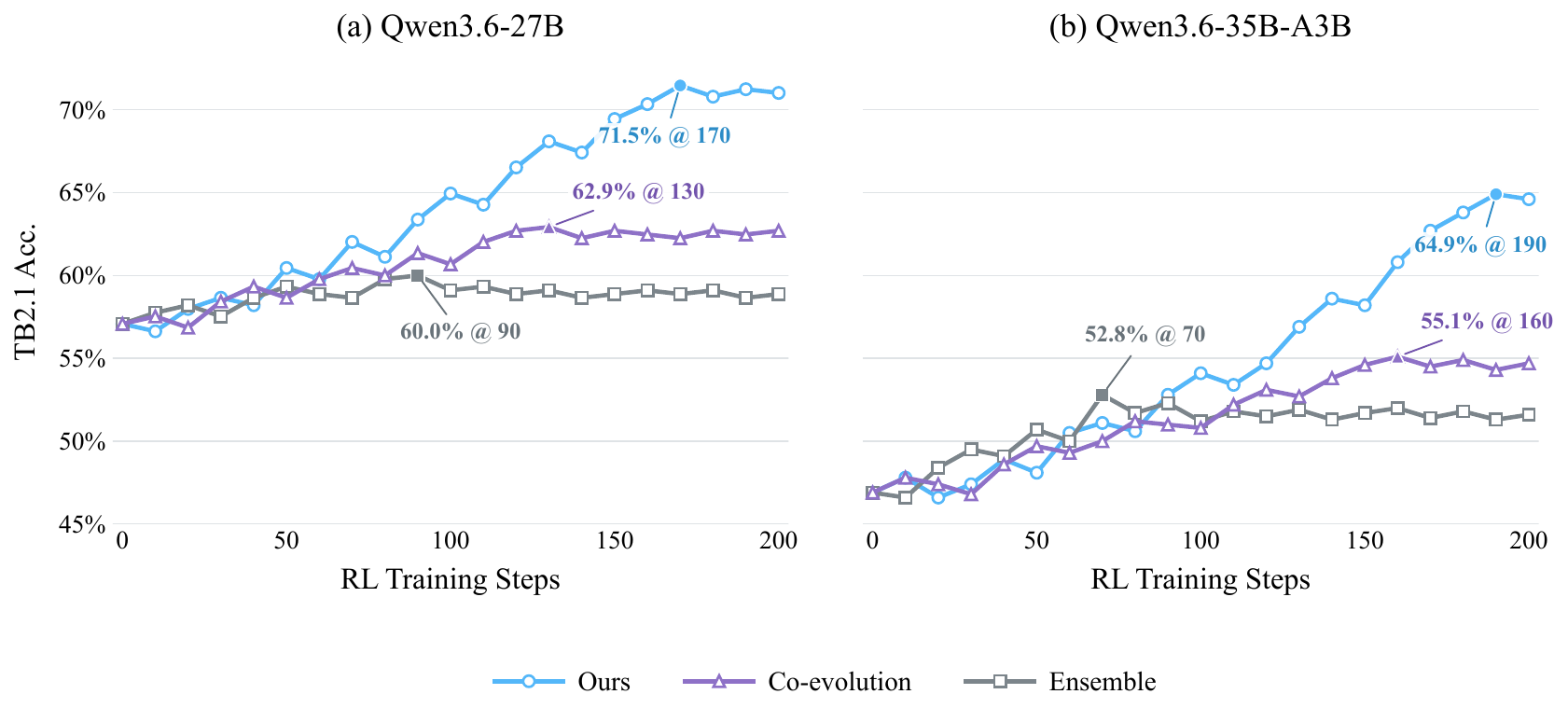}
    \caption{Checkpoint performance for Qwen3.6-27B (left) and Qwen3.6-35B-A3B (right) over 200 RL training steps, with offline evaluation every 10 steps; annotations mark each method's peak accuracy and corresponding training step. Step 0 denotes the shared checkpoint after RFT used to increase policy entropy before RL training.}
    \label{fig:tb21-training-dynamics}
\end{figure}

\paragraph{Environment Ensemble.}
The target agent is first evaluated on the seed environments, and the environments it fails to solve are selected.
All two- and three-environment combinations are then enumerated, with each combination merged into one harder environment.
This procedure is applied recursively as a tree, with the outputs of one level becoming the inputs to the next.

\paragraph{Agent--Environment Co-Evolution.}
The target agent first rolls out on the seed environments to generate trajectories.
Its failed trajectories are analyzed to construct a weakness bank, from which combinations are sampled to synthesize new, harder environments.
After the agent trains for 50 RL steps, the procedure is repeated.

\paragraph{Environment Evolution.}
Environment Evolution randomly orders the three modes at each generation and follows the cross-mode fallback, producing a verified 15-generation lineage from each seed without target-agent rollouts.

Figure~\ref{fig:tb21-training-dynamics} reports offline Terminal-Bench~2.1 evaluation every 10 RL steps.
Environment evolution reaches peak accuracies of 71.5\% and 64.9\% on Qwen3.6-27B and Qwen3.6-35B-A3B, compared with 62.9\% and 55.1\% for Co-evolution and 60.0\% and 52.8\% for Ensemble.

\section{Conclusion}
In this paper, we propose \emph{environment evolution}, which incrementally synthesizes environments of increasing difficulty for RL training.
It serves as a more general paradigm for scaling environments for terminal agents, built on the derived off-policy formulation of environment difficulty.
Across different models, evolved environments show a consistent increase in difficulty generation by generation despite being synthesized off-policy.
By scheduling these environments along lineages, long-horizon RL training on Qwen3.6-27B and Qwen3.6-35B-A3B receives continuously effective learning signals and achieves higher performance than co-evolution and ensemble baselines.
Future work will explore the potential of environment evolution for SWE agents and Computer-Use Agents.

\bibliography{custom}

@misc{yang2024sweagentagentcomputerinterfacesenable,
      title={SWE-agent: Agent-Computer Interfaces Enable Automated Software Engineering}, 
      author={John Yang and Carlos E. Jimenez and Alexander Wettig and Kilian Lieret and Shunyu Yao and Karthik Narasimhan and Ofir Press},
      year={2024},
      eprint={2405.15793},
      archivePrefix={arXiv},
      primaryClass={cs.SE},
      url={https://arxiv.org/abs/2405.15793}, 
}

@misc{merrill2026terminalbenchbenchmarkingagentshard,
      title={Terminal-Bench: Benchmarking Agents on Hard, Realistic Tasks in Command Line Interfaces}, 
      author={Mike A. Merrill and Alexander G. Shaw and Nicholas Carlini and Boxuan Li and Harsh Raj and Ivan Bercovich and Lin Shi and Jeong Yeon Shin and Thomas Walshe and E. Kelly Buchanan and Junhong Shen and Guanghao Ye and Haowei Lin and Jason Poulos and Maoyu Wang and Marianna Nezhurina and Jenia Jitsev and Di Lu and Orfeas Menis Mastromichalakis and Zhiwei Xu and Zizhao Chen and Yue Liu and Robert Zhang and Leon Liangyu Chen and Anurag Kashyap and Jan-Lucas Uslu and Jeffrey Li and Jianbo Wu and Minghao Yan and Song Bian and Vedang Sharma and Ke Sun and Steven Dillmann and Akshay Anand and Andrew Lanpouthakoun and Bardia Koopah and Changran Hu and Etash Guha and Gabriel H. S. Dreiman and Jiacheng Zhu and Karl Krauth and Li Zhong and Niklas Muennighoff and Robert Amanfu and Shangyin Tan and Shreyas Pimpalgaonkar and Tushar Aggarwal and Xiangning Lin and Xin Lan and Xuandong Zhao and Yiqing Liang and Yuanli Wang and Zilong Wang and Changzhi Zhou and David Heineman and Hange Liu and Harsh Trivedi and John Yang and Junhong Lin and Manish Shetty and Michael Yang and Nabil Omi and Negin Raoof and Shanda Li and Terry Yue Zhuo and Wuwei Lin and Yiwei Dai and Yuxin Wang and Wenhao Chai and Shang Zhou and Dariush Wahdany and Ziyu She and Jiaming Hu and Zhikang Dong and Yuxuan Zhu and Sasha Cui and Ahson Saiyed and Arinbjörn Kolbeinsson and Jesse Hu and Christopher Michael Rytting and Ryan Marten and Yixin Wang and Alex Dimakis and Andy Konwinski and Ludwig Schmidt},
      year={2026},
      eprint={2601.11868},
      archivePrefix={arXiv},
      primaryClass={cs.SE},
      url={https://arxiv.org/abs/2601.11868}, 
}

@misc{chen2021evaluatinglargelanguagemodels,
      title={Evaluating Large Language Models Trained on Code}, 
      author={Mark Chen and Jerry Tworek and Heewoo Jun and Qiming Yuan and Henrique Ponde de Oliveira Pinto and Jared Kaplan and Harri Edwards and Yuri Burda and Nicholas Joseph and Greg Brockman and Alex Ray and Raul Puri and Gretchen Krueger and Michael Petrov and Heidy Khlaaf and Girish Sastry and Pamela Mishkin and Brooke Chan and Scott Gray and Nick Ryder and Mikhail Pavlov and Alethea Power and Lukasz Kaiser and Mohammad Bavarian and Clemens Winter and Philippe Tillet and Felipe Petroski Such and Dave Cummings and Matthias Plappert and Fotios Chantzis and Elizabeth Barnes and Ariel Herbert-Voss and William Hebgen Guss and Alex Nichol and Alex Paino and Nikolas Tezak and Jie Tang and Igor Babuschkin and Suchir Balaji and Shantanu Jain and William Saunders and Christopher Hesse and Andrew N. Carr and Jan Leike and Josh Achiam and Vedant Misra and Evan Morikawa and Alec Radford and Matthew Knight and Miles Brundage and Mira Murati and Katie Mayer and Peter Welinder and Bob McGrew and Dario Amodei and Sam McCandlish and Ilya Sutskever and Wojciech Zaremba},
      year={2021},
      eprint={2107.03374},
      archivePrefix={arXiv},
      primaryClass={cs.LG},
      url={https://arxiv.org/abs/2107.03374}, 
}

@misc{lee2026metaharnessendtoendoptimizationmodel,
      title={Meta-Harness: End-to-End Optimization of Model Harnesses}, 
      author={Yoonho Lee and Roshen Nair and Qizheng Zhang and Kangwook Lee and Omar Khattab and Chelsea Finn},
      year={2026},
      eprint={2603.28052},
      archivePrefix={arXiv},
      primaryClass={cs.AI},
      url={https://arxiv.org/abs/2603.28052}, 
}

@misc{wang2025openhandsopenplatformai,
      title={OpenHands: An Open Platform for AI Software Developers as Generalist Agents}, 
      author={Xingyao Wang and Boxuan Li and Yufan Song and Frank F. Xu and Xiangru Tang and Mingchen Zhuge and Jiayi Pan and Yueqi Song and Bowen Li and Jaskirat Singh and Hoang H. Tran and Fuqiang Li and Ren Ma and Mingzhang Zheng and Bill Qian and Yanjun Shao and Niklas Muennighoff and Yizhe Zhang and Binyuan Hui and Junyang Lin and Robert Brennan and Hao Peng and Heng Ji and Graham Neubig},
      year={2025},
      eprint={2407.16741},
      archivePrefix={arXiv},
      primaryClass={cs.SE},
      url={https://arxiv.org/abs/2407.16741}, 
}

@misc{yao2023reactsynergizingreasoningacting,
      title={ReAct: Synergizing Reasoning and Acting in Language Models}, 
      author={Shunyu Yao and Jeffrey Zhao and Dian Yu and Nan Du and Izhak Shafran and Karthik Narasimhan and Yuan Cao},
      year={2023},
      eprint={2210.03629},
      archivePrefix={arXiv},
      primaryClass={cs.CL},
      url={https://arxiv.org/abs/2210.03629}, 
}

@misc{gandhi2026endlessterminalsscalingrl,
      title={Endless Terminals: Scaling RL Environments for Terminal Agents}, 
      author={Kanishk Gandhi and Shivam Garg and Noah D. Goodman and Dimitris Papailiopoulos},
      year={2026},
      eprint={2601.16443},
      archivePrefix={arXiv},
      primaryClass={cs.LG},
      url={https://arxiv.org/abs/2601.16443}, 
}

@misc{wu2026largescaleterminalagentictrajectory,
      title={Large-Scale Terminal Agentic Trajectory Generation from Dockerized Environments}, 
      author={Siwei Wu and Yizhi Li and Yuyang Song and Wei Zhang and Yang Wang and Riza Batista-Navarro and Xian Yang and Mingjie Tang and Bryan Dai and Jian Yang and Chenghua Lin},
      year={2026},
      eprint={2602.01244},
      archivePrefix={arXiv},
      primaryClass={cs.CL},
      url={https://arxiv.org/abs/2602.01244}, 
}

@misc{fan2026scalableterminaltasksynthesis,
      title={Toward Scalable Terminal Task Synthesis via Skill Graphs}, 
      author={Zhiyuan Fan and Tinghao Yu and Yuanjun Cai and Jiangtao Guan and Yun Yang and Dingxin Hu and Jiang Zhou and Xing Wu and Zhuo Han and Feng Zhang and Lilin Wang},
      year={2026},
      eprint={2604.25727},
      archivePrefix={arXiv},
      primaryClass={cs.AI},
      url={https://arxiv.org/abs/2604.25727}, 
}

@misc{bercovich2026makesgoodterminalagentbenchmark,
      title={What Makes a Good Terminal-Agent Benchmark Task: A Guideline for Adversarial, Difficult, and Legible Evaluation Design}, 
      author={Ivan Bercovich},
      year={2026},
      eprint={2604.28093},
      archivePrefix={arXiv},
      primaryClass={cs.AI},
      url={https://arxiv.org/abs/2604.28093}, 
}

@misc{bui2026buildingeffectiveaicoding,
      title={Building Effective AI Coding Agents for the Terminal: Scaffolding, Harness, Context Engineering, and Lessons Learned}, 
      author={Nghi D. Q. Bui},
      year={2026},
      eprint={2603.05344},
      archivePrefix={arXiv},
      primaryClass={cs.AI},
      url={https://arxiv.org/abs/2603.05344}, 
}

@misc{pi2026dataengineeringscalingllm,
      title={On Data Engineering for Scaling LLM Terminal Capabilities}, 
      author={Renjie Pi and Grace Lam and Mohammad Shoeybi and Pooya Jannaty and Bryan Catanzaro and Wei Ping},
      year={2026},
      eprint={2602.21193},
      archivePrefix={arXiv},
      primaryClass={cs.CL},
      url={https://arxiv.org/abs/2602.21193}, 
}

@misc{hua2026cliuniverseverifiabletasksynthesis,
      title={{CLI-Universe}: Towards Verifiable Task Synthesis Engine for Terminal Agents},
      author={Zhanbo Hua and Yifan Yao and Weihao Xie and Yongchi Zhao and Minghao Liu and Ruizhi Qiu and Zhewei Huang and Zun Wang and Yiyan Ji and Yunhai Ye and Letian Zhu and Xinping Lei and Han Li and Zhiyuan Ma and Zili Wang and Zhaoxiang Zhang and Jiaheng Liu},
      year={2026},
      eprint={2606.22883},
      archivePrefix={arXiv},
      primaryClass={cs.AI},
      url={https://arxiv.org/abs/2606.22883},
}

@misc{zhao2026nexforge,
      title={{NexForge}: Scaling Agent Capabilities through Requirement-Driven Task Synthesis for {LLMs}},
      author={Jiarong Zhao and Zhikai Lei and Zhiheng Xi and Rui Zheng and Hang Yan and Jie Zhou and Qin Chen and Liang He},
      year={2026},
      eprint={2607.14186},
      archivePrefix={arXiv},
      primaryClass={cs.SE},
      url={https://arxiv.org/abs/2607.14186},
}

@misc{yao2026learninggeneralizablebehaviorsterminal,
      title={Learning Generalizable Behaviors for Terminal Agents},
      author={Yihang Yao and Bo Pang and Xuan Phi Nguyen and Ding Zhao and Shafiq Joty and Semih Yavuz},
      year={2026},
      eprint={2608.22631},
      archivePrefix={arXiv},
      primaryClass={cs.LG},
      url={https://arxiv.org/abs/2608.22631},
}

@misc{wang2019pairedopenendedtrailblazerpoet,
      title={Paired Open-Ended Trailblazer (POET): Endlessly Generating Increasingly Complex and Diverse Learning Environments and Their Solutions}, 
      author={Rui Wang and Joel Lehman and Jeff Clune and Kenneth O. Stanley},
      year={2019},
      eprint={1901.01753},
      archivePrefix={arXiv},
      primaryClass={cs.NE},
      url={https://arxiv.org/abs/1901.01753}, 
}

@misc{wang2020enhancedpoetopenendedreinforcement,
      title={Enhanced POET: Open-Ended Reinforcement Learning through Unbounded Invention of Learning Challenges and their Solutions}, 
      author={Rui Wang and Joel Lehman and Aditya Rawal and Jiale Zhi and Yulun Li and Jeff Clune and Kenneth O. Stanley},
      year={2020},
      eprint={2003.08536},
      archivePrefix={arXiv},
      primaryClass={cs.NE},
      url={https://arxiv.org/abs/2003.08536}, 
}

@misc{openendedlearningteam2021openendedlearningleadsgenerally,
      title={Open-Ended Learning Leads to Generally Capable Agents}, 
      author={Open Ended Learning Team and Adam Stooke and Anuj Mahajan and Catarina Barros and Charlie Deck and Jakob Bauer and Jakub Sygnowski and Maja Trebacz and Max Jaderberg and Michael Mathieu and Nat McAleese and Nathalie Bradley-Schmieg and Nathaniel Wong and Nicolas Porcel and Roberta Raileanu and Steph Hughes-Fitt and Valentin Dalibard and Wojciech Marian Czarnecki},
      year={2021},
      eprint={2107.12808},
      archivePrefix={arXiv},
      primaryClass={cs.LG},
      url={https://arxiv.org/abs/2107.12808}, 
}

@misc{dennis2021emergentcomplexityzeroshottransfer,
      title={Emergent Complexity and Zero-shot Transfer via Unsupervised Environment Design}, 
      author={Michael Dennis and Natasha Jaques and Eugene Vinitsky and Alexandre Bayen and Stuart Russell and Andrew Critch and Sergey Levine},
      year={2021},
      eprint={2012.02096},
      archivePrefix={arXiv},
      primaryClass={cs.LG},
      url={https://arxiv.org/abs/2012.02096}, 
}

@misc{jiang2021prioritizedlevelreplay,
      title={Prioritized Level Replay}, 
      author={Minqi Jiang and Edward Grefenstette and Tim Rocktäschel},
      year={2021},
      eprint={2010.03934},
      archivePrefix={arXiv},
      primaryClass={cs.LG},
      url={https://arxiv.org/abs/2010.03934}, 
}

@misc{jiang2022replayguidedadversarialenvironmentdesign,
      title={Replay-Guided Adversarial Environment Design}, 
      author={Minqi Jiang and Michael Dennis and Jack Parker-Holder and Jakob Foerster and Edward Grefenstette and Tim Rocktäschel},
      year={2022},
      eprint={2110.02439},
      archivePrefix={arXiv},
      primaryClass={cs.LG},
      url={https://arxiv.org/abs/2110.02439}, 
}

@misc{parkerholder2023evolvingcurricularegretbasedenvironment,
      title={Evolving Curricula with Regret-Based Environment Design}, 
      author={Jack Parker-Holder and Minqi Jiang and Michael Dennis and Mikayel Samvelyan and Jakob Foerster and Edward Grefenstette and Tim Rocktäschel},
      year={2023},
      eprint={2203.01302},
      archivePrefix={arXiv},
      primaryClass={cs.LG},
      url={https://arxiv.org/abs/2203.01302}, 
}

@misc{zala2024envgengeneratingadaptingenvironments,
      title={EnvGen: Generating and Adapting Environments via LLMs for Training Embodied Agents}, 
      author={Abhay Zala and Jaemin Cho and Han Lin and Jaehong Yoon and Mohit Bansal},
      year={2024},
      eprint={2403.12014},
      archivePrefix={arXiv},
      primaryClass={cs.CL},
      url={https://arxiv.org/abs/2403.12014}, 
}

@misc{hu2025agentgenenhancingplanningabilities,
      title={AgentGen: Enhancing Planning Abilities for Large Language Model based Agent via Environment and Task Generation}, 
      author={Mengkang Hu and Pu Zhao and Can Xu and Qingfeng Sun and Jianguang Lou and Qingwei Lin and Ping Luo and Saravan Rajmohan},
      year={2025},
      eprint={2408.00764},
      archivePrefix={arXiv},
      primaryClass={cs.CL},
      url={https://arxiv.org/abs/2408.00764}, 
}

@inproceedings{qi2025webrltrainingllmweb,
      title={WebRL: Training LLM Web Agents via Self-Evolving Online Curriculum Reinforcement Learning}, 
      author={Zehan Qi and Xiao Liu and Iat Long Iong and Hanyu Lai and Xueqiao Sun and Jiadai Sun and Xinyue Yang and Yu Yang and Shuntian Yao and Wei Xu and Jie Tang and Yuxiao Dong},
      booktitle={The Thirteenth International Conference on Learning Representations},
      year={2025},
      url={https://proceedings.iclr.cc/paper_files/paper/2025/hash/c66e1fcc9691aae706250638f36f681b-Abstract-Conference.html},
}

@misc{guo2025genenvdifficultyalignedcoevolutionllm,
      title={GenEnv: Difficulty-Aligned Co-Evolution Between LLM Agents and Environment Simulators}, 
      author={Jiacheng Guo and Ling Yang and Peter Chen and Qixin Xiao and Yinjie Wang and Xinzhe Juan and Jiahao Qiu and Ke Shen and Mengdi Wang},
      year={2025},
      eprint={2512.19682},
      archivePrefix={arXiv},
      primaryClass={cs.CL},
      url={https://arxiv.org/abs/2512.19682}, 
}

@inproceedings{chen2025scalingagentlearning,
      title={Scaling Agent Learning via Experience Synthesis},
      author={Zhaorun Chen and Zhuokai Zhao and Kai Zhang and Bo Liu and Qi Qi and Yifan Wu and Tarun Kalluri and Sara Cao and Yuanhao Xiong and Haibo Tong and Huaxiu Yao and Hengduo Li and Jiacheng Zhu and Xian Li and Dawn Song and Bo Li and Jason Weston and Dat Huynh},
      booktitle={The Fourteenth International Conference on Learning Representations},
      year={2026},
      url={https://openreview.net/forum?id=cf7qpBwttr},
}

@inproceedings{yang2026coevolve,
      title={{CoEvolve}: Training {LLM} Agents via Agent-Data Mutual Evolution},
      author={Shidong Yang and Ziyu Ma and Tongwen Huang and Yiming Hu and Yong Wang and Xiangxiang Chu},
      booktitle={Proceedings of the 64th Annual Meeting of the Association for Computational Linguistics},
      pages={23015--23036},
      year={2026},
      doi={10.18653/v1/2026.acl-long.1055},
      url={https://aclanthology.org/2026.acl-long.1055/},
}

@misc{liu2026spade,
      title={{SPADE}: Self-Play in Adaptive Synthetic Executable Environments},
      author={Bo Liu and Simon Yu and Yiding Jiang and Ao Qu and Andrew Zhao and Zichen Liu and Junsu Kim and Zijian Zhou and Seungone Kim and Tongzheng Ren and Mickel Liu and Hanfei Yu and Zhaorun Chen and Weiyan Shi and Paul Pu Liang and Luke Zettlemoyer and Yejin Choi and Natasha Jaques},
      year={2026},
      eprint={2608.19197},
      archivePrefix={arXiv},
      primaryClass={cs.CL},
      url={https://arxiv.org/abs/2608.19197},
}

@misc{sygkounas2026covolveadversarialcoevolutionlargelanguagemodelgenerated,
      title={COvolve: Adversarial Co-Evolution of Large-Language-Model-Generated Policies and Environments via Two-Player Zero-Sum Game}, 
      author={Alkis Sygkounas and Rishi Hazra and Andreas Persson and Pedro Zuidberg Dos Martires and Amy Loutfi},
      year={2026},
      eprint={2603.28386},
      archivePrefix={arXiv},
      primaryClass={cs.AI},
      url={https://arxiv.org/abs/2603.28386}, 
}

@article{sutton1999between,
  title={Between MDPs and semi-MDPs: A framework for temporal abstraction in reinforcement learning},
  author={Sutton, Richard S. and Precup, Doina and Singh, Satinder},
  journal={Artificial Intelligence},
  volume={112},
  number={1--2},
  pages={181--211},
  year={1999},
  publisher={Elsevier},
  doi={10.1016/S0004-3702(99)00052-1}
}

@article{kaelbling1998planning,
  title={Planning and acting in partially observable stochastic domains},
  author={Kaelbling, Leslie Pack and Littman, Michael L and Cassandra, Anthony R},
  journal={Artificial Intelligence},
  volume={101},
  number={1-2},
  pages={99--134},
  year={1998},
  publisher={Elsevier}
}

@misc{fu2026areallargescaleasynchronousreinforcement,
      title={AReaL: A Large-Scale Asynchronous Reinforcement Learning System for Language Reasoning}, 
      author={Wei Fu and Jiaxuan Gao and Xujie Shen and Chen Zhu and Zhiyu Mei and Chuyi He and Shusheng Xu and Guo Wei and Jun Mei and Jiashu Wang and Tongkai Yang and Binhang Yuan and Yi Wu},
      year={2026},
      eprint={2505.24298},
      archivePrefix={arXiv},
      primaryClass={cs.LG},
      url={https://arxiv.org/abs/2505.24298}, 
}

@misc{bellemare2013arcade,
      title={The Arcade Learning Environment: An Evaluation Platform for General Agents}, 
      author={Marc G. Bellemare and Yavar Naddaf and Joel Veness and Michael Bowling},
      year={2013},
      eprint={1207.4708},
      archivePrefix={arXiv},
      primaryClass={cs.AI},
      doi={https://doi.org/10.1613/jair.3912},
      url={https://arxiv.org/abs/1207.4708}, 
}

@misc{brockman2016openai,
      title={OpenAI Gym}, 
      author={Greg Brockman and Vicki Cheung and Ludwig Pettersson and Jonas Schneider and John Schulman and Jie Tang and Wojciech Zaremba},
      year={2016},
      eprint={1606.01540},
      archivePrefix={arXiv},
      primaryClass={cs.LG},
      url={https://arxiv.org/abs/1606.01540}, 
}

@misc{skyrlagent,
      title={SkyRL-Agent: Efficient RL Training for Multi-turn LLM Agent}, 
      author={Shiyi Cao and Dacheng Li and Fangzhou Zhao and Shuo Yuan and Sumanth R. Hegde and Connor Chen and Charlie Ruan and Tyler Griggs and Shu Liu and Eric Tang and Richard Liaw and Philipp Moritz and Matei Zaharia and Joseph E. Gonzalez and Ion Stoica},
      year={2025},
      eprint={2511.16108},
      archivePrefix={arXiv},
      primaryClass={cs.AI},
      url={https://arxiv.org/abs/2511.16108}, 
}

@misc{earl,
      title={EARL: Efficient Agentic Reinforcement Learning Systems for Large Language Models}, 
      author={Zheyue Tan and Mustapha Abdullahi and Tuo Shi and Huining Yuan and Zelai Xu and Chao Yu and Boxun Li and Bo Zhao},
      year={2025},
      eprint={2510.05943},
      archivePrefix={arXiv},
      primaryClass={cs.DC},
      url={https://arxiv.org/abs/2510.05943}, 
}

@misc{cobbe2019procgen,
      title={Leveraging Procedural Generation to Benchmark Reinforcement Learning}, 
      author={Karl Cobbe and Christopher Hesse and Jacob Hilton and John Schulman},
      year={2020},
      eprint={1912.01588},
      archivePrefix={arXiv},
      primaryClass={cs.LG},
      url={https://arxiv.org/abs/1912.01588}, 
}

@misc{garcin2024dred,
      title={DRED: Zero-Shot Transfer in Reinforcement Learning via Data-Regularised Environment Design}, 
      author={Samuel Garcin and James Doran and Shangmin Guo and Christopher G. Lucas and Stefano V. Albrecht},
      year={2024},
      eprint={2402.03479},
      archivePrefix={arXiv},
      primaryClass={cs.LG},
      url={https://arxiv.org/abs/2402.03479}, 
}

@misc{shao2024deepseekmathpushinglimitsmathematical,
      title={DeepSeekMath: Pushing the Limits of Mathematical Reasoning in Open Language Models}, 
      author={Zhihong Shao and Peiyi Wang and Qihao Zhu and Runxin Xu and Junxiao Song and Xiao Bi and Haowei Zhang and Mingchuan Zhang and Y. K. Li and Y. Wu and Daya Guo},
      year={2024},
      eprint={2402.03300},
      archivePrefix={arXiv},
      primaryClass={cs.CL},
      url={https://arxiv.org/abs/2402.03300}, 
}

@misc{touvron2023llama2,
      title={Llama 2: Open Foundation and Fine-Tuned Chat Models},
      author={Hugo Touvron and others},
      year={2023},
      eprint={2307.09288},
      archivePrefix={arXiv},
      primaryClass={cs.CL},
      url={https://arxiv.org/abs/2307.09288},
}

@misc{tencent2026hy4preview,
      title={{Hy4 preview}},
      author={{Tencent Hunyuan}},
      year={2026},
      month=aug,
      howpublished={\url{https://hy.tencent.com/research/hy4-preview}},
}

@misc{anthropic2025claudecode,
      title={{Claude Code}},
      author={{Anthropic}},
      year={2025},
      howpublished={\url{https://code.claude.com/docs/en/overview}},
}

@misc{ma2025stabilizingmoereinforcement,
      title={Stabilizing {MoE} Reinforcement Learning by Aligning Training and Inference Routers},
      author={Wenhan Ma and Hailin Zhang and Liang Zhao and Yifan Song and Yudong Wang and Zhifang Sui and Fuli Luo},
      year={2025},
      eprint={2510.11370},
      archivePrefix={arXiv},
      primaryClass={cs.CL},
      url={https://arxiv.org/abs/2510.11370},
}
\bibliographystyle{iclr2026_conference}

\appendix
\section{Version Note}
We will provide more details in future versions.
Future versions will also report RSI results in which the same model both constructs and learns from the environments.

\end{document}